\documentclass[sigconf,screen]{acmart}

\AtBeginDocument{%
  \providecommand\BibTeX{{%
    \normalfont B\kern-0.5em{\scshape i\kern-0.25em b}\kern-0.8em\TeX}}}

\usepackage{gensymb}
\usepackage{algorithmic}
\usepackage[linesnumbered, ruled]{algorithm2e}
\usepackage{comment}
\usepackage{subfig}
\usepackage{paralist}
\usepackage{tabularx}
\usepackage{xcolor}
\usepackage{pgfplotstable}
\usepackage{booktabs}
\usepackage{microtype}

\pgfplotsset{compat=1.17}

\copyrightyear{2021}
\acmYear{2021}
\setcopyright{acmcopyright}
\acmConference[ISSTA '21]{Proceedings of the 30th ACM SIGSOFT International Symposium on Software Testing and Analysis}{July 11--17, 2021}{Virtual, Denmark}
\acmBooktitle{Proceedings of the 30th ACM SIGSOFT International Symposium on Software Testing and Analysis (ISSTA '21), July 11--17, 2021, Virtual, Denmark}
\acmPrice{15.00}
\acmDOI{10.1145/3460319.3464802}
\acmISBN{978-1-4503-8459-9/21/07}

\begin{document}

\title{Automatic Test Suite Generation for Key-Points Detection DNNs using Many-Objective Search (Experience Paper)}

\author{Fitash Ul Haq}
\orcid{0000-0003-2253-9085}
\affiliation{%
  \institution{University of Luxembourg}
  \country{Luxembourg}
}
\email{fitash.ulhaq@uni.lu}

\author{Donghwan Shin}
\orcid{0000-0002-0840-6449}
\affiliation{%
  \institution{University of Luxembourg}
  \country{Luxembourg}
}
\email{donghwan.shin@uni.lu}

\author{Lionel C. Briand}
\orcid{0000-0002-1393-1010}
\affiliation{%
  \institution{University of Luxembourg}
  \country{Luxembourg}
}
\affiliation{%
  \institution{University of Ottawa}
  \city{Ottawa}
  \country{Canada}
}
\email{lionel.briand@uni.lu}

\author{Thomas Stifter}
\affiliation{%
  \institution{IEE S.A.}
  \country{Luxembourg}
}
\email{thomas.stifter@iee.lu}

\author{Jun Wang}
\orcid{0000-0002-0481-5341}
\affiliation{%
  \institution{Post Luxembourg}
  \country{Luxembourg}
}
\email{junwang.lu@gmail.com}

\begin{abstract}
Automatically detecting the positions of key-points (e.g., facial key-points or finger key-points) in an image is an essential problem in many applications, such as driver's gaze detection and drowsiness detection in automated driving systems. With the recent advances of Deep Neural Networks (DNNs), Key-Points detection DNNs (KP-DNNs) have been increasingly employed for that purpose. Nevertheless, KP-DNN testing and validation have remained a challenging problem because KP-DNNs predict many independent key-points at the same time---where each individual key-point may be critical in the targeted application---and images can vary a great deal according to many factors. 

In this paper, we present an approach to automatically generate test data for KP-DNNs using many-objective search. In our experiments, focused on facial key-points detection DNNs developed for an industrial automotive application, we show that our approach can generate test suites to severely mispredict, on average, more than 93\% of all key-points. In comparison, random search-based test data generation can only severely mispredict 41\% of them. Many of these mispredictions, however, are not avoidable and should not therefore be considered failures. We also empirically compare state-of-the-art, many-objective search algorithms and their variants, tailored for test suite generation. Furthermore, we investigate and demonstrate how to learn specific conditions, based on image characteristics (e.g., head posture and skin color), that lead to severe mispredictions. Such conditions serve as a basis for risk analysis or DNN retraining.
\end{abstract}



\maketitle

\section{Introduction}\label{sec:intro}

Automatically detecting key-points (e.g., facial key-points or finger
key-points) in an image or a video is a fundamental step for many
applications, such as face recognition~\cite{facialrecog}, facial
expression recognition~\cite{facialExpressionRecog}, and drowsiness
detection~\cite{jabbar2018real}. With the recent advances in Deep
Neural Networks (DNNs), Key-Points detection DNNs (KP-DNNs) have been
widely studied~\cite{huang2017local,huang2015deepfinger,jabbar2018real,song2020kpnet}. 

To ensure the reliability of KP-DNNs, it is essential to check how
accurate the DNNs are when applied to various test data. Nevertheless,
testing KP-DNNs is still often focused on pre-recorded test data, such
as publicly available
datasets~\cite{wolf2011face,andriluka20142d,sapp2013modec} that are
typically not collected or generated according to any systematic
strategy, and therefore provide limited test results and
confidence. Further, since 3rd parties with ML expertise often
provide such KP-DNNs, the internal information of the KP-DNNs is usually not
available to the engineers who integrate them into their application
systems~\cite{lin2020black,papernot2017practical}. As a result,
black-box testing approaches should be favored, preferably one that is
dedicated and adapted to KP-DNNs to maximize test effectiveness within
acceptable time constraints, which is the objective of this paper. 

To automatically generate new and diverse test data, one may opt to
use advanced DNN testing approaches that have been recently proposed
by several researchers. For example, \citet{DeepTest} presented
DeepTest, an approach that generates new test images from training
images by applying simple image transformations (e.g., rotate, scale,
and for and rain effects). \citet{wicker2018feature} generated
adversarial examples, i.e., small perturbations that are almost
imperceptible by humans but causing DNN misclassifications, using
feature extraction from images. \citet{asfault} presented AsFault, an
approach that generates virtual road networks in computer simulations
using search-based testing for testing a DNN-based automated driving
system. However, none of the existing DNN testing approaches take into
account the following testing challenges that are specific to KP-DNNs
predicting multiple key-points at the same time. First, accuracy for
individual key-points is of great importance, and therefore one should
not simply consider average prediction errors across all key-points.
Second, the number of key-points is typically large, e.g., our
evaluation uses a facial KP-DNN that detects 27 key-points in an
image. Third, depending on the performance of the KP-DNN under test,
it may be infeasible to generate a test image causing a significant
misprediction for some particular key-points. Therefore, if such
infeasibility is observed at run time, it is essential to dynamically
and efficiently distribute the computational resources dedicated to
testing to the other key-points. This implies that using one of the
existing DNN testing approaches is not an option. It also prevents us
from running an independent search for each individual key-point. 

To address the above challenges, we recast the problem of KP-DNN
testing as a many-objective optimization problem. Since the severe
misprediction of each individual key-point becomes one objective,
many-objective optimization algorithms can potentially be effective
and efficient at generating test data that causes the KP-DNN to
mispredict individual key-points. Though our approach is applicable to
any KP-DNNs, we focus our experiments on a particular but common 
industrial application, Facial KP-DNNs (FKP-DNN).
The experimental subjects are provided by our industry 
partner IEE in the automotive domain, who is working on driver 
gaze detection systems that either warn the driver or take the control 
of the vehicle when the driver is apparently not paying attention 
on the road while driving. Our empirical investigation
shows that our approach is effective at generating test data that
cause most of the key-points (93\% on average) to be severely
mispredicted. Though many of these mispredictions are not avoidable
because a key-point can be invisible due to, for example, a shadow,
characterizing them is important to analyze risks and a large number
of them can still be addressed by retraining or other means.  We
further investigate (1) the degree of mispredictions caused by test
data generated by alternative search algorithms and (2) how to learn
specific conditions (e.g., head posture) leading to severe
mispredictions of individual key-points. Our ultimate goal is to
provide practical recommendations on how to test KP-DNNs and how to
analyze test results to support risk analysis and retraining. 

The contributions of this paper are summarized as follows:
\begin{compactenum}[1)]
    \item We formalize the problem definition of KP-DNN testing. Our formalism specifies KP-DNN input and output variables, as well as the notion of severe misprediction for a key-point, and characterises the specific testing challenges in this context.
    \item We propose a way to automatically generate test data for KP-DNNs using many-objective search algorithms and simulation.
    \item We investigate how automatically generated test data and results can be effectively used to reveal and possibly address the root causes of inaccuracy in FKP-DNNs under test. 
    \item We present empirical results and lessons learned drawn from our experience in applying the approach in an industrial context. 
\end{compactenum}
    
The rest of the paper is organized as follows.
Section~\ref{sec:background} provides background information on
KP-DNNs and search-based testing with simulation.
Section~\ref{sec:problem} formalises the problem of KP-DNN testing.
Section~\ref{related-work} positions our work with respect to related
work. Section~\ref{sec:approach} describes our approach.
Section~\ref{sec:expr} evaluates our approach with an industrial case
study and summarizes lessons learned.
Section~\ref{sec:conclusion} concludes the paper.
\section{Background} \label{sec:background}

\subsection{Search-based Testing}\label{sec:sbst} 
Search-based software testing (SBST)~\cite{sbst} uses
meta-heuristic algorithms to automate software testing tasks, such
as test case generation~\cite{fraser2011evosuite} and
prioritization~\cite{li2007search}, for a specific system under test.
The key idea is to formulate a software testing problem as an
optimization problem by defining proper fitness functions. For
example, EvoSuite~\cite{fraser2011evosuite} uses a search-based
approach to automatically generate unit test cases for a Java program
to satisfy a coverage criterion, such as branch coverage. In this
case, the fitness function is defined based on the coverage achieved
by unit test cases.

Recently, SBST has also been used for DNN
testing. For example, AsFault~\cite{asfault} automatically
generates virtual roads to make vision-based DNNs go out of lane, and
DeepJanus~\cite{deepjanus} uses multi-objective search to
generate a pair of similar test inputs that cause the DNN under 
test to misbehave for one test input but not for the other.

However, when the number of objectives is above three, multi-objective
search algorithms, such as NSGA-II~\cite{nsga-ii}, do not scale
well~\cite{knowles2007quantifying,ciro2016nsga}. This is where
many-objective search algorithms come into play. For example,
NSGA-III~\cite{nsga-iii} is a generic many-objective search algorithm that
extends NSGA-II with the idea of virtual reference points to increase
the diversity of optimal solutions even when there are more than three
objectives. 
Panichella et al.~\cite{mosa} proposed a Many-Objective Sorting
Algorithm (MOSA), another extension of NSGA-II, that is tailored for
test suite generation. In contrast to NSGA-III, MOSA aims to
efficiently achieve each objective individually. To do this, MOSA has
three main features: (1) it focuses search towards uncovered objectives,
(2) it uses a novel preference criterion to rank solutions rather than
diversifying them, and (3) it saves the best test case for each
objective in an archive. MOSA has shown to outperform alternative search 
algorithms in the context of coverage-based unit testing for traditional 
software programs~\cite{7102604,mosa}.
Recently, Abdessalem et al.~\cite{fitest} proposed FITEST, an
extension of MOSA, to further improve the efficiency of many-objective
search. The idea is to dynamically reduce the number of candidates
(i.e., the population size) to be considered by focusing on the
uncovered objectives only. Hence, FITEST's population size decreases
as more objectives are achieved, whereas MOSA's population size is
fixed throughout the search. 

Notice that MOSA and FITEST are carefully designed for search-based
test suite generation when the number of objectives is above three.
Therefore, they are a priori suitable to address the problem of test suite
generation for KPF-DNNs, as detailed in Section~\ref{sec:problem}.
\subsection{Simulation-based Testing} 
Testing cyber-physical, safety-critical systems can be done in two
ways: (1) in a real-world environment and (2) in a virtual
environment, relying on simulators. In the former case, software is
deployed in the real-world environment, a form of testing which is
 often expensive and dangerous. The latter case, referred to as
simulation-based testing, can raise concerns about simulation's
fidelity but has two major benefits. First, simulation-based testing
offers controllability, meaning that test drivers can control static
and dynamic features of the simulation (e.g., light, face features),
and can thus automatically cover a wide range of scenarios. Second,
because we can control simulation parameters and can get image
information from the simulator, we know the ground truth (actual FKPs)
and can thus apply search-based solutions to perform safe and fully
automated testing of safety-critical
software~\cite{testing-vision,Testing-autonomous-cars,testing-adas}.

In many cyber-physical fields, simulation models are developed before implementing
actual products. These simulation models help engineers in various
activities, such as early verification and validation of the system
under test~\cite{Approximation-Refinement-Testing,hou2015simulation}.
In such contexts, simulated-based testing is highly recommended as it
is economical, faster, safer and flexible~\cite{hou2015simulation}.
\section{Problem Definition }\label{sec:problem}
In this section, we provide a general but precise problem description regarding test
data generation for FKP-DNNs. Though we use a specific application domain for the purpose of exemplification, this description can easily be
generalized to all situations where key-points must be detected in an
image, such as hand key-point detection \cite{simon2017hand} and 
human pose key-points detection~\cite{hourglass}, regardless of the content of that image.

\begin{figure}
	\centering
	\includegraphics[width=0.6\linewidth]{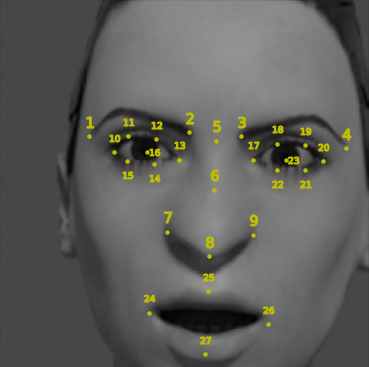}
    \caption{The actual positions of 27 facial key-points}
    \label{fig:27-keypoints}
\end{figure}

A FKP-DNN takes as input a facial image (real or simulated) and
returns the predicted positions of its key-points;
Figure~\ref{fig:27-keypoints} shows an example image with the
actual positions of key-points. An image can be 
defined by various factors, such as face size,
skin color, and head posture. Using a labeled image that contains the
actual positions of key-points, one can measure the degree of
prediction errors for an FKP-DNN under test. The goal of test suite
generation for FKP-DNNs is to generate labeled images that cause an
FKP-DNN under test to inaccurately predict the positions of key-points, 
to an extent which affects the safety of the system relying on such
predictions.

More specifically, let $t$ be a labeled test image composed of a tuple
$t = (\mathbf{ic}_t, \mathbf{p}_t)$ where $\mathbf{ic}_t$ represents image
characteristics, such as face size, skin color, and head posture, and
$\mathbf{p}_t$ represents the actual positions of key-points in $t$.
$\mathbf{p}_t$ can be further decomposed as $\mathbf{p}_t = \langle p_{t,1},
\dots, p_{t,k} \rangle$ where $p_{t,i}$ for $i=1,\dots,k$ is the actual
position of the $i$-th key-point in $t$ and $k$ is the total number of
key-points. Depending on $\mathbf{ic}_t$, some key-points may not actually be
visible in $t$. For example, if the head is turned 90$\degree$ to the
right, the right eye key-point will be invisible in
the image taken from the front camera. In this case, the actual positions of
invisible key-points are \textit{null}. $V(t)$ is a set of indices for
visible key-points in $t$, and $|V(t)|$ is the total number of
visible key-points in $t$. An FKP-DNN $d$ can be considered as a
function $d(t) = \langle \hat{p}_{t,1}, \dots, \hat{p}_{t,k} \rangle$ where
$\hat{p}_{t,i}$ is the predicted position of the $i$-th key-point in $t$.
The Normalized Mean Error (NME) of $d$ for all key-points in $t$
can be defined as follows:
\begin{displaymath}
\mathit{NME}(d, t) = \frac{\Sigma_{i\in V(t)} \mathit{NE}(p_{t,i},\hat{p}_{t,i})}{|V(t)|}
\end{displaymath}
where $\mathit{NE}(p_{t,i}, \hat{p}_{t,i})$ is the Normalized Error of
the predicted position $\hat{p}_{t,i}$ with respect to the actual
position $p_{t,i}$, ranging between 0 and 1. 
For example, in our case study, if one measures
the error using the Euclidean distance between the actual and
predicted positions in an image, then the normalization can be done by
dividing the distance by the height or width of the face\footnote{This
can be easily calculated with the actual positions of key-points.},
whichever is larger. Such a normalization allows error values to be
compared for faces of different sizes.

Generating a critical test image $t$ for $d$ such that it maximizes
$\mathit{NME}(d, t)$ could be the goal of test case generation.
However, it does not properly account for the fact that accuracy for
individual key-points is of great importance, and therefore that
averaging prediction errors across key-points may be misleading. This
is, indeed, a major concern for our industry partner IEE, who is developing DNN-based gaze detection
systems that either warn the driver or take the control of the vehicle
when the driver is not paying attention while driving. By
considering individual key-points, we may observe that certain
key-points' predictions are particularly inaccurate. Such
information can be used by IEE to (1) focus the retraining of $d$
on these key-points for improving its accuracy or (2) select accurate 
key-points to be primarily relied upon---when there are alternative 
choices---by the applications using them to make decisions.

To verify if $d$ correctly predicts individual key-points for $t$, we
need to check if $\mathit{NE}(p_{t,i}, \hat{p}_{t,i})$ is less than a
small threshold $\epsilon$ for all key-points. Therefore, 
though that may not be possible, test suite generation aims to
generate a minimal set of test cases $\mathit{TS}$ that satisfies
$\mathit{NE}(p_{t,i}, \hat{p}_{t,i}) \ge \epsilon$ for all
$i=1,\dots,k$ for some $t\in \mathit{TS}$. Note that the value of
$\epsilon$ is usually application specific; for example, IEE uses
$\epsilon = 0.05$ because $\mathit{NME} > 0.05$ is considered critical
in their applications.

Test suite generation for FKP-DNNs entails several challenges. First,
the test input space is too large to be exhaustively explored. For
IEE, this is clear when considering the features that characterize a
facial image, including head posture and facial characteristics, such
as different shapes of eyes, noses, and mouths. Second, the
number of key-points is typically large (e.g., 27 key-points for IEE's
DNN) and, increasing the complexity of the problem, 
individual key-points are independent as people may exhibit
different relative positions for the same key-points. 
Third, depending on the
accuracy of the DNN under test, it may be infeasible to find an image
causing the normalized error to exceed the threshold for some
key-points. Considering a limited time test budget, if finding a
critical image for a key-point seems infeasible, it is essential to
dynamically and efficiently distribute computation time at run time to
find critical images for the other key-points. Fourth, it is not easy
to manually label the actual positions of key-points for a large
number of test images. While there are publicly available real-world
datasets for various key-point detection
problems~\cite{wolf2011face,andriluka20142d,sapp2013modec}, such
datasets limit the exploration of the input space to what data is
available since there are no publicly available simulators to generate additional images. 
Last but not least, FKP-DNNs, like other DNNs in practice,
are often provided by third parties with expertise in ML, who
typically do not provide access to the trained DNN's internal
information. Therefore, test
suite generation for FKP-DNNs should be black-box, in the sense that
it should not rely on DNN internal information.

To address the above challenges, as described in detail in
Section~\ref{sec:approach}, we suggest to apply many-objective
search algorithms, which can be effective and efficient for
achieving many independent objectives within a limited time budget. 
These algorithms are a priori a good match to our problem since 
requirements for individual key-points 
(i.e., $\mathit{NE}(p_{t,i},\hat{p}_{t,i}) \ge \epsilon$ for 
$t\in \mathit{TS}$) can be translated into objective functions.
Furthermore, this approach is DNN-agnostic as it considers 
the DNN under test as a black-box. 
\section{Related Work}\label{related-work}

Many automated techniques are available in the literature for testing DNNs.
One type of automatic test data generation techniques is to generate adversarial examples~\cite{yuan2019adversarial} that are imperceptible for humans but cause the DNN under test to misbehave.
\citet{guo2018dlfuzz} proposed DLFuzz, an approach that iteratively applies small perturbations to original images to maximize neuron coverage (i.e., the rate of activated neurons) and prediction differences between the original and synthesized images.
\citet{deepbillboard} and \citet{phygan} proposed DeepBillboard and PhysGAN, respectively, two similar approaches that generate adversarial images that can be placed on drive-by billboards, both digitally or physically, to induce failures in DNN-based automated driving systems.
\citet{wicker2018feature} presented an approach to generate adversarial images using feature extraction techniques, such as Scale Invariant Feature Transform (SIFT), to extract features from original images. 
\citet{rozsa2019facial} introduced a Fast Flipping Attribute (FFA) technique to effectively generate adversarial images for DNNs detecting facial attributes (e.g., male or female).
However, in contrast to our objective of verifying the safety of DNN predictions for a large variety of plausible test inputs, adversarial examples mainly target security attacks where an attacker intentionally introduces human-imperceptible changes (i.e., attacks) to cause the DNN to generate incorrect predictions. 

Another important line of work is to generate new test data from already labeled data (e.g., training data) by applying label-preserving changes to avoid labeling problems for newly generated test inputs.
\citet{DeepExplore} proposed DeepXplore, an approach that generates label-preserving test images to maximize both neuron coverage and differential behaviors of multiple DNNs for the generated images.
\citet{DeepTest} introduced DeepTest, an approach that synthesizes label-preserving test images by applying affine transformations and effect filters, such as rain and snow, to original images in order to maximize neuron coverage.
\citet{DeepRoad} presented DeepRoad, an approach that uses Generative Adversarial Networks (GANs), instead of simple transformations and effect filters, to generate more realistic images.
\citet{du2018deepcruiser} proposed DeepCruiser, an approach that generates label-preserving sequences of test data to test stateful deep learning systems, based on Recurrent Neural Networks (RNNs) using special coverage criteria for RNNs converted to Markov Decision Process (MDP).
Recently, \citet{xie2019deephunter} presented DeepHunter, an extensible coverage-guided testing framework extending DeepTest to further utilize multiple coverage criteria, with more label-preserving transformation strategies.
While these works enable engineers to generate more realistic test data from existing data, generating label-preserving test images inherently limits the search space when the objective is to identify critical situations in the most comprehensive way possible. Furthermore, when relying on a simulator, labeling is not a critical issue. 

To overcome the limitation of label-preserving test data generation, simulation-based testing is increasingly used for testing DNNs, especially in the context of DNN-based automated driving systems.
\citet{asfault} presented AsFault, a search-based approach to generates different types of road topologies in a simulated environment to test the lane keeping functionality of self-driving DNNs.
\citet{tuncali2018simulation} introduced Sim-ATAV, a framework for testing closed-loop behaviors of DNN-based systems in simulated environments using requirement falsification methods.
\citet{deepjanus} presented DeepJanus, an approach to generate pairs of similar test inputs, using search-based testing with simulations, that cause the DNN under test to misbehave for one test input but not for the other.
Although the solutions in this category successfully used simulation to generate critical test data, they are inadequate for the test suite generation of FKP-DNNs, because they do not consider many independent outputs individually. Therefore, even when it seems infeasible to find a critical image for a certain key-point, these solutions cannot dynamically redistribute computational resources to to other key-points.

In summary, existing work does not address the fundamental and specific challenges of test suite generation for KP-DNNs, as described in Section~\ref{sec:problem}. To this end, we propose a solution, based on many-objective optimization, that automatically generates test suites whose objective is to cause KP-DNNs to severely mispredict the positions of individual key-points and then use machine learning to explain such mispredictions.
\section{Search-based Test Suite Generation}\label{sec:approach}

This section provides a solution to the problem of test suite 
generation for KP-DNNs, described in Section~\ref{sec:problem}, by
applying many-objective search algorithms. 
Similar to Section~\ref{sec:problem}, although our approach is not 
specific to the domain of facial key-points, to root the presentation 
into concrete examples, we describe the approach in the context of 
test suite generation for FKP-DNNs.

Recall that MOSA~\cite{mosa} and FITEST~\cite{fitest} take as input
(1) a set of objectives, (2) a set of fitness functions indicating
the degree to which individual objectives have been achieved, and (3)
a time budget; it then returns a set of solutions that maximally
achieve each objective individually within the given time budget.
Therefore, we can apply the algorithms to our problem by carefully
defining a set of corresponding objectives and fitness functions.

Based on the problem definition in Section~\ref{sec:problem}, we can
define that an objective for each key-point is to find a test
image that makes a FKP-DNN under test \textit{severely mispredict} the
key-point position. Specifically, for a FKP-DNN
$d$, the set of objectives to be achieved by a many-objective search
algorithm is $\mathit{NE}(p_{t, i}, \hat{p}_{t, i}) \ge \epsilon$
for all $i=1,\dots,k$ where $\mathit{NE}(p_{t, i}, \hat{p}_{t, i})$
is the normalized error of the predicted position $\hat{p}_{t, i}$
with respect to the actual position $p_{t, i}$ for the $i$-th
key-point and $\epsilon$ is a small threshold pre-defined by domain
experts. Naturally, the set of fitness functions corresponding to the
objectives is $\mathit{NE}(p_{t, i}, \hat{p}_{t, i})$ for all
$i=1,\dots,k$.

While it seems intuitive to define the sets of objectives and fitness
functions for many-objective search algorithms, the issue in
practice is how to determine the actual positions of key-points in an
automated manner; a simulator plays a key role here.

\begin{figure}
	\centering
	\includegraphics[width=\linewidth]{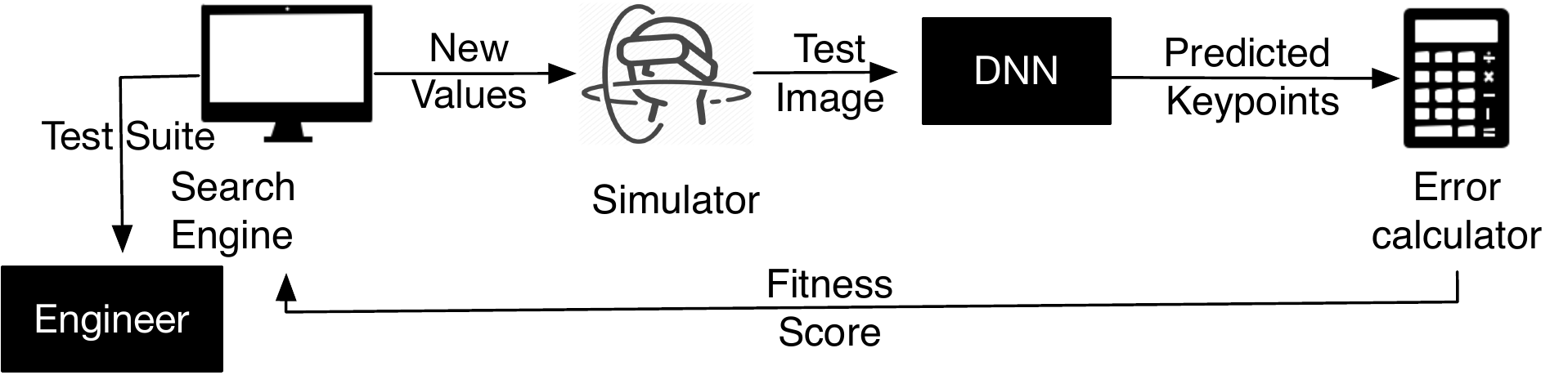}
	\caption{Overview of automatic test suite generation for FKP-DNNs}
	\label{fig:approach}
\end{figure}

Figure \ref{fig:approach} shows the overview of the
search-based test suite generation process for a FKP-DNN using a
simulator. It is composed of four main components: a search engine, a
simulator, the DNN under test, and a fitness calculator. The process
begins with the search engine generating a set of new input values for
image characteristics $\mathbf{ic}$ (e.g., head posture and skin color). 
The input values are used by the simulator to generate a new test image
$t$ as well as the actual positions of key-points $\mathbf{p}_t =
\langle p_{t,1}, \dots, p_{t,k} \rangle$ for all $i=1,\dots,k$. The
FKP-DNN under test $d$ takes $t$ as input and returns the predicted
positions of the key-points $d(t) = \langle \hat{p}_{t,1}, \dots,
\hat{p}_{t,k} \rangle$. Then, the fitness calculator computes the
fitness score of $t$ using $\mathit{NE}(p_{t, i}, \hat{p}_{t, i})$ for
each $i=1,\dots,k$. The fitness score is fed back to the search engine
to generate new and better input values over the next iterations. This
process continues until either a given time budget runs out or all the
objectives are achieved. The process ends with returning a test suite
$\mathit{TS}$ such that each $t\in \mathit{TS}$ satisfies
$\mathit{NE}(p_{t, i}, \hat{p}_{t, i}) \ge \epsilon$ for some $i=1,\dots,k$. 
In the following subsections, we will explain each of the approach components 
in more detail.

\subsection{Search Engine}\label{sec:search-engine}
The search engine drives the whole process based on a meta-heuristic
search algorithm. For a given set of objectives with fitness
functions and a time budget, it iteratively generates new test inputs
for image characteristics ($\mathbf{ic}$) to ultimately provide engineers with
the most effective test suite. Search algorithms, such as Random Search (RS),
MOSA~\cite{mosa}, and FITEST~\cite{fitest}, vary in the way they
generate new test inputs based on the fitness results from previous
iterations.

For example, MOSA starts with an initial set of randomly generated
test cases that forms an initial population; then, it creates new test
cases using \textit{crossover} and \textit{mutation} operators that are
typically used in Genetic Algorithms (GA) to find better candidates 
while promoting their diversity. Unlike GA,
the \textit{selection} in MOSA is performed by considering both the
non-dominance relation and the uncovered objectives. Specifically, for
each uncovered objective $u$, a test case $t$ will have a higher chance of
remaining in the next generation if $t$ is non-dominated by the
others and is the closest to cover $u$. To form the final test suite,
MOSA uses an \textit{archive} that keeps track of the best test cases
that cover individual objectives. FITEST is basically the same as
MOSA, except that it dynamically reduces the population size with each
iteration, according to the number of uncovered objectives.

\begin{algorithm}
\SetKwInOut{Input}{Input}
\SetKwInOut{Output}{Output}
\small

\Input{Set of Objectives $O$}
\Output{Archive (Test Cases) $A$}
    Archive $A \gets \emptyset$ \label{alg:mo:arc} \\
    Set of Uncovered Objectives $U \gets O$ \label{alg:mo:inituc} \\
    Set of Test Cases $P \gets \mathit{initialPopulation}(|O|)$ \label{alg:mo:init}\\ 
    Set of Covered Objectives $C \gets \mathit{calculateObjs}(O, P)$ \label{alg:mo:calobjs1}\\
    $A \gets \mathit{updateArchive}(A, P, C)$ \label{alg:mo:upac1}\\
    $U \gets \mathit{updateUncoveredObjs}(U, C)$\label{alg:mo:upobj1}
    
    \While{$\mathit{not (stoping\_condition)}$}{
        Set of Test Cases $Q \gets \mathit{generateOffspring}(P)$ \label{alg:mo:offsp}\\
        $C \gets \mathit{calculateObjs}(O, Q)$ \label{alg:mo:calobj2}\\
        $A \gets \mathit{updateArchive} (A, Q, C)$\label{alg:mo:uparc2}\\
        $U \gets \mathit{updateUncoveredObjs(U, C)}$\label{alg:mo:upobj2}\\
        $P \gets \mathit{getNextGeneration}(P \cup Q, U)$ \label{alg:mo:ngeration}\\
    }
    \textbf{return} $A$
\caption{Pseudo-code of MOSA and FITEST}
\label{alg:mo}
\end{algorithm}

More specifically, Algorithm~\ref{alg:mo} presents the pseudo-code 
of MOSA and FITEST. It takes as input a set of objectives $O$ and 
returns an archive $A$ (i.e., a test suite) that aims to maximally 
achieve individual objectives in $O$. To do that, the algorithm 
begins with initializing $A$, the set of uncovered objectives $U$, 
and the population $P$ 
(lines~\ref{alg:mo:arc}--\ref{alg:mo:init}). The algorithm then 
computes the set of covered objectives $C\subseteq O$ by calculating 
the fitness scores of $P$ for $O$ (line~\ref{alg:mo:calobjs1}), 
updates $A$ to include test cases from $P$, which are best at 
achieving $C$ (line~\ref{alg:mo:upac1}), and updates $U$ to remove 
the covered objectives in $C$ (line~\ref{alg:mo:upobj1}). Until the 
stopping criterion is met, the algorithm repeats the followings: 
(1) generating a set of new test cases $Q$ from $P$ using 
\textit{crossover} and \textit{mutation} (line~\ref{alg:mo:offsp}), 
(2) updating $C$, $A$, and $U$ for $Q$, as done for $P$ 
(lines~\ref{alg:mo:calobj2}--\ref{alg:mo:upobj2}), and (3) generating 
the next generation $P$ from the current $P$ and $Q$ considering 
$U$ using \textit{selection} (line~\ref{alg:mo:ngeration}). 
The algorithm ends by returning $A$. The main difference between 
MOSA and FITEST is in the \textit{getNextGeneration} function 
(line~\ref{alg:mo:ngeration}): MOSA keeps $|P| = |O|$, whereas 
FITEST reduces $|P|$ as $|U|$ decreases.

To improve the performance of MOSA and FITEST, we can also consider their 
variants, namely MOSA+ and FITEST+. These variants are
identical to the originals, but the crossover strategy is changed to
only consider fitness scores for uncovered (not-yet-achieved)
objectives to better guide new test cases towards them.
In other words, the higher the fitness score for uncovered objectives, the
more similar children are to parents. This is done by dynamically
updating the distribution index of the Simulated Binary Crossover
(SBX), based on the fitness score of parents, for uncovered objectives.

In practice, using an effective and efficient search algorithm, for a
given problem, is critical for testing at scale and therefore be
applicable in industrial contexts. This is also the case here for
generating better test suites for FKP-DNNs.

\subsection{Simulator}\label{sec:sim}
The simulator is one of the key enablers of our approach because it takes as
input $\mathbf{ic}_t$ and generates $t$ \emph{labeled} with $\mathbf{p}_t$. It
should be able to generate diverse test images by manipulating various
$\mathbf{ic}$s, such that there is variation in  head posture, gender, skin
color, and mouth size. At the same time, it should be able to generate
individual test images within reasonable time since this strongly affects the
efficiency of the entire search-based process.

To achieve this, one can use 3D modeling tools, such as
MakeHuman~\cite{makehuman} and Blender~\cite{blender}. MakeHuman can be used to
create parameterized 3D face models with detailed morphological characteristics,
such as skin color and hair style. Blender, on the other hand, can be used to
manipulate the 3D models according to selected $\mathbf{ic}$
values~\cite{10.100723}. Since a 3D model is basically a textured polygonal mesh
(i.e., a collection of nodes and edges in 3D) that defines the shape of a
polyhedral object, an engineer can mark the actual positions of key-points in
the model by selecting the corresponding nodes in the textured mesh. Then, using
the simulator, it is easy to generate a 2D image capturing the face labeled with
the actual key-point positions.

\subsection{Fitness Calculator and Fitness Functions}
The fitness calculator takes $p_{t, i}$ and $\hat{p}_{t, i}$
and returns $\mathit{NE}(p_{t, i}, \hat{p}_{t, i})$ for
each $i=1,\dots,k$. As mentioned in Section~\ref{sec:problem}, one way
to calculate $\mathit{NE}(p_{t, i}, \hat{p}_{t, i})$ is to compute the
Euclidean distance between $p_{t, i}$ and $\hat{p}_{t, i}$ and then
normalize it using the height or width of the actual face, whichever
is larger. Normalization enables the comparison of the error values 
of different 3D models having different face sizes.

\section{Empirical Evaluation}\label{sec:expr}
In this section, we report on the empirical evaluation of our many-objective,
search-based test suite generation approach when applied to an industrial FKP-DNN developed by
IEE. We aim to answer the following research questions.

\textit{\textbf{RQ1:} How do alternative many-objective search
algorithms fare in terms of test effectiveness?} RQ1 aims to
check whether using many-objective search algorithms, such as MOSA
and FITEST, is indeed a suitable solution for the problem of test
suite generation for FKP-DNNs. To answer this, we investigate how
many key-points are severely mispredicted (according to our earlier
definition) by test suites that are automatically generated by the
search algorithms, and how alternative algorithms compare to each
other and to a simple random search.

\textit{\textbf{RQ2}: Can we further distinguish search algorithms
using the degree of mispredictions caused by the test suites they
generate?} Since RQ1 only considers the number of severely
mispredicted key-points, differences in effectiveness across search
algorithms may not appear clearly and completely. For example, two test
suites generated by different algorithms may cause the same number of
severely mispredicted key-points. However, since the level of risk
entailed by mispredictions may be proportional to the distance
between the actual and predicted key-points, a test suite with a
higher degree of mispredictions may be more amenable to risk analysis, depending on
the application context. Based on this, RQ2 compares how
\textit{severely} key-points are mispredicted by test suites
generated across different search algorithms.

\textit{\textbf{RQ3}: Can we explain individual key-point
mispredictions in terms of image characteristics?}
In addition to the cost-effective, automated testing of FKP-DNNs,
engineers need to understand why severe mispredictions occur for
individual key-points. This is critical for engineers to analyze and
address the root causes of mispredictions, or, when the latter is not
possible, at the very least assess the risks. For example, if
engineers know that a certain key-point is significantly mispredicted
for a specific head posture range, they can generate more test images
in that range and retrain the DNN to improve its prediction accuracy.
To this end, RQ3 aims to investigate whether it is possible to provide
accurate and interpretable explanations of mispredictions based on
image characteristics used by the simulator to generate test images.

\subsection{Case Study Design}
We use a proprietary FKP-DNN, denoted IEE-DNN (described in detail in
\S~\ref{sec:IEE-DNN}), developed by IEE to build a driver's gaze
detection component for autonomous vehicles. To enable
simulation-based testing, we also use an in-house simulator, namely
IEE-SIM (described in detail in \S~\ref{sec:IEE-SIM}), developed by
IEE to generate large numbers of labeled facial images to train the IEE-DNN.

\subsubsection{DNN}\label{sec:IEE-DNN}
IEE-DNN takes as input a 256x256 pixel image; it returns the
predicted positions of 27 facial key-points in the input image.
Figure~\ref{fig:27-keypoints} depicts the actual positions of the 27
key-points for a sample image.

The IEE-DNN is based on the stacked hourglass architecture~\cite{hourglass}
with an adaptive wing loss function~\cite{wang2019adaptive}.
Specifically, it consists of two hourglass modules, each of which
contains four residual modules for downsampling and another four for upsampling.
It is trained on 18,120 syntactic images generated by the IEE-SIM.
Against an independent set of 2,738 syntactic test images,
the IEE-DNN achieved a low NME value of $0.018$, on average.
Given that the threshold for labelling mispredictions as severe for
individual key-points is $0.05$, such NME value implies that,
if we just consider the average prediction error for all key-points,
the IEE-DNN is sufficiently accurate according to the test set.
Note that our approach is independent from any specific DNN architecture
as it does not require any DNN internal information.

\subsubsection{Simulator}\label{sec:IEE-SIM}
IEE-SIM takes as input various image characteristics, such as head
posture, light intensity, and the position of a (virtual) camera, and
returns a corresponding facial image with the actual positions of the
27 key-points.

The IEE-SIM is based on MakeHuman~\cite{makehuman} and
Blender~\cite{blender}. To decrease its execution time, IEE engineers
carefully designed many 3D face models in advance, by considering the diversity
in skin colors and the shapes and sizes of faces, mouths, and noses. By doing
this, we can quickly generate an image by specifying a pre-defined 3D model ID
to use instead of dynamically generating the 3D models from scratch at run time.
As a result, the average execution time for generating one labeled image is
around 6 seconds on an iMac (3GHz 6-Core Intel i5 CPU,  40GB memory, and Radeon
Pro 570X 4GB graphic card).

The main concern of IEE is to verify the accuracy
of the IEE-DNN for diverse head postures and drivers.
Therefore, we manipulate four feature values when varying image
characteristics: roll, pitch, and yaw values for controlling head
posture and the 3D model ID for indirectly the face features to be
used. The ranges of the roll, pitch, and yaw values are limited
between -30$\degree$ and +30$\degree$ to mimic realistic ways in which drivers
position their head. For 3D models, we use the subset of 10 different
models that IEE considered to be of main interest.

\subsection{RQ1: Effectiveness of Test Suites}

\subsubsection{Setup}\label{sec:rq1-setup}
To answer RQ1, we generate a test suite using our approach for a fixed
time budget and measure the \textit{Effectiveness Score} (\textit{ES})
of the test suite, defined as the proportion of key-points that are
severely mispredicted by the IEE-DNN according to the test suite over the total
number of key-points. \textit{ES} ranges between 0 and 1, where higher
values, when possible, are desirable.

To better understand how \textit{ES} varies across different search
algorithms, we compare MOSA~\cite{mosa}, FITEST~\cite{fitest}, and their
variants MOSA+ and FITEST+ (see Section~\ref{sec:search-engine}).
We use Random Search (RS) as a baseline. RS randomly
generates a test suite for each iteration and keeps the best until the
search process ends; RS provides insights into how easy the search
problem is and helps us assess if other, more complex search
algorithms are indeed necessary.

One might consider using NSGA-III as another baseline that tries to
achieve many objectives collectively. However, it is not readily
tailored to test suite generation as explained in
Section~\ref{sec:sbst},  and therefore requires addressing the problem
of how to encode test suites into chromosomes, which is a research
subject that is beyond the scope of our paper.

For MOSA, FITEST, and their variants, the initial population size is the
number of objectives. To be consistent, we set the number of randomly
generated test cases at each iteration of RS to be the
number of objectives. For the other parameters, such as mutation and
crossover rates in MOSA, FITEST, and their variants, we use the
default values recommended in the original studies.

To account for randomness in search algorithms, we repeat the
experiment 20 times. For each run, we use the same time budget of two
hours for all search algorithms, based on our preliminary evaluation
showing that two hours are enough to converge. We apply the non-parametric
Mann–Whitney U test~\cite{mann1947test} to assess the statistical significance of
the difference in \textit{ES} between algorithms.
We also measure Vargha and Delaney's $\hat{A}_{AB}$~\cite{5002101}
to capture the effect size of the difference.

\subsubsection{Results}\label{sec:rq1-results}
\begin{figure}
	\centering
	\includegraphics[width=0.8\linewidth]{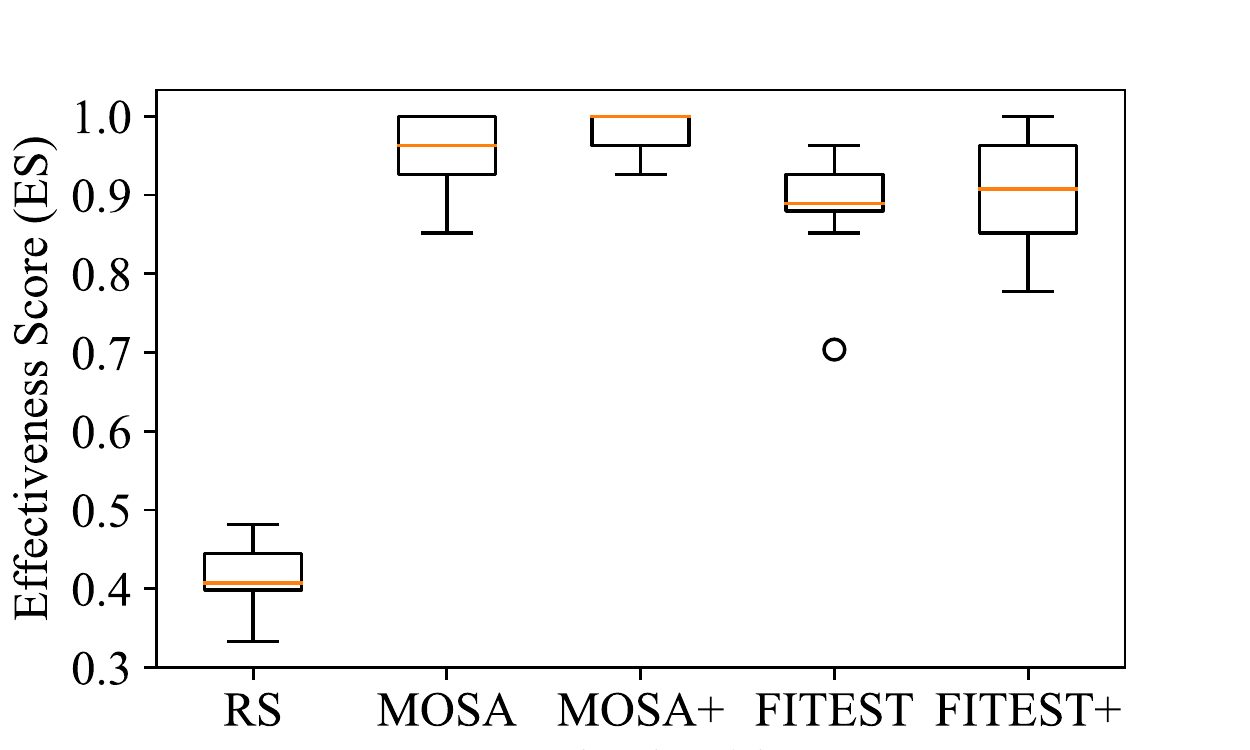}
	\caption{Test effectiveness for different search algorithms}
	\label{fig:comparison-10-runs}
\end{figure}

Figure~\ref{fig:comparison-10-runs} depicts the differences in
\textit{ES} across search algorithms. RS only achieves $\mathit{ES} =
0.41$ on average across 20 runs. Though such $\mathit{ES}$ value may
seem a priori high, one must recall that it cannot be interpreted as a
failure rate in the normal sense since certain key-points are
impossible to predict, e.g., some key-points are hidden by a shadow.
In general, it is to be expected that such DNNs will have inherent
limitations that cannot be addressed by retraining or any other means.
It is up to the system using a DNN to adequately address such
limitations, for example by not using key-point predictions in
shadowed areas and relying on the other key-points for decisions.
Therefore, the purpose of $\textit{ES}$ values is only to compare the
test effectiveness of different algorithms and not to assess the
accuracy of the DNN under test. In contrast to RS, MOSA, MOSA+, and
FITEST+ reach $\mathit{ES} = 1$ in at least one run, meaning that
these algorithms could generate test suites that cause the IEE-DNN to
severely mispredict all 27 key-points. Across 20 runs, the
many-objective search algorithms achieve $\mathit{ES} > 0.93$ on
average, implying that our approach seems indeed effective at
generating test suites that cause severe mispredictions for many
key-points.

Table~\ref{table:statistics} shows the results of statistical
comparisons between different search algorithms. Under the
\textit{Pair} column, sub-columns \textit{A} and \textit{B} indicate
the two search algorithms being compared. Under the \textit{ES}
column, sub-columns \textit{$p$-value} and $\hat{A}_{AB}$ indicate the
statistical significance and the effect size, respectively, when
    comparing \textit{ES} distributions between \textit{A} and \textit{B}.

\begin{table}
\caption{Statistical Analysis Results}
\label{table:statistics}
\centering
\begin{filecontents*}{data/pvalues.csv}
A,B,ES-Pvalues,ES-Effectsize,MS-Pvalues,MS-Effectsize
MOSA,RS,0,1,0,0.897
MOSA+,RS,0,1,0,0.902
FITEST,RS,0,1,0,0.876
FITEST+,RS,0,1,0,0.873
MOSA,FITEST,0,0.837,0.57,0.541
MOSA+,FITEST,0,0.945,0.594,0.543
FITEST+,FITEST,0.7502,0.53,0.052,0.507
MOSA,FITEST+,0.0045,0.7575,0.177,0.545
MOSA+,FITEST+,0,0.86,0.009,0.556
MOSA+,MOSA,0.1091,0.6375,0.78,0.5
\end{filecontents*}
\pgfplotstabletypeset[
	col sep=comma,
	every head row/.style={before row={
        \toprule
        \multicolumn{2}{c}{Pair}
        & \multicolumn{2}{c}{\textit{ES}}
        & \multicolumn{2}{c}{\textit{MS}} \\
        \cmidrule(lr){1-2} \cmidrule(lr){3-4} \cmidrule(lr){5-6}
    }, after row=\midrule},
	every last row/.style={after row=\bottomrule},
    fixed,fixed zerofill,empty cells with={-},
    columns/A/.style={string type,column type=l,column name=A},
	columns/B/.style={string type,column type=l,column name=B},
	columns/ES-Pvalues/.style={column type=r,column name=$p$-value,precision=3},
	columns/ES-Effectsize/.style={column type=r,column name=$\hat{A}_{AB}$,precision=2},
	columns/MS-Pvalues/.style={column type=r,column name=$p$-value,precision=3},
	columns/MS-Effectsize/.style={column type=r,column name=$\hat{A}_{AB}$,precision=2},
]{data/pvalues.csv}
\end{table}

At a level of significance $\alpha$ = 0.01, we can see that MOSA is
significantly better than FITEST, with large effect size (0.84).
Comparing MOSA+ and FITEST+ shows the same result, with an even larger
effect size (0.86). This implies that (dynamically) reducing the
population size at each iteration significantly degrades the
effectiveness of the search algorithms for test suite generation, as it
decreases their ability to explore the search space.
When comparing MOSA (and FITEST) and MOSA+ (and FITEST+), their
differences in \textit{ES} are statistically insignificant, implying
that dynamically controlling the similarity between parents and
children in crossover does not significantly improve effectiveness. In
RQ2, we will further compare the many-objective search algorithms by
considering the degree of misprediction severity across the test suites
they generate.

To summarize, the answer to RQ1 is that our approach is effective in
generating test suites that cause IEE-DNN---which is already rather
accurate (NME = $0.018$ for the test dataset) as one might expect from
an industrial model---to severely mispredict more than 93\% of all
key-points on average. This number is also much higher than that
obtained with random search (41\%). While MOSA and MOSA+ are
significantly better than FITEST and FITEST+ in terms of \textit{ES},
there is no significant difference between MOSA and MOSA+, which will
be investigated further in RQ2.

\subsection{RQ2: Misprediction Severity for Individual Key-Points}

\subsubsection{Setup}\label{sec:rq2-setup}
To answer RQ2, following the same procedure as for RQ1, we measure the
\textit{Misprediction Severity} (\textit{MS}) of a test suite for
each key-point, defined as the maximum $\mathit{NE}$ value observed when
running the test suite. As for $\mathit{NE}$, \textit{MS}
ranges between 0 and 1, where 1 implies the maximum prediction error.

Since RQ2 aims to further distinguish the search algorithms by
considering the \textit{MS} values of test suites they generated in RQ1,
we use the test suites from RQ1 and report the average \textit{MS} of
individual key-points for each algorithm. We apply the
non-parametric Wilcoxon signed-rank test to statistically compare \textit{MS}
distributions for each key-point between search algorithms, and measure
Vargha and Delaney's $\hat{A}_{AB}$ to capture the effect size of the
difference.

\subsubsection{Results}\label{sec:rq2-results}
\begin{figure}
	\centering
	\includegraphics[width=0.8\linewidth]{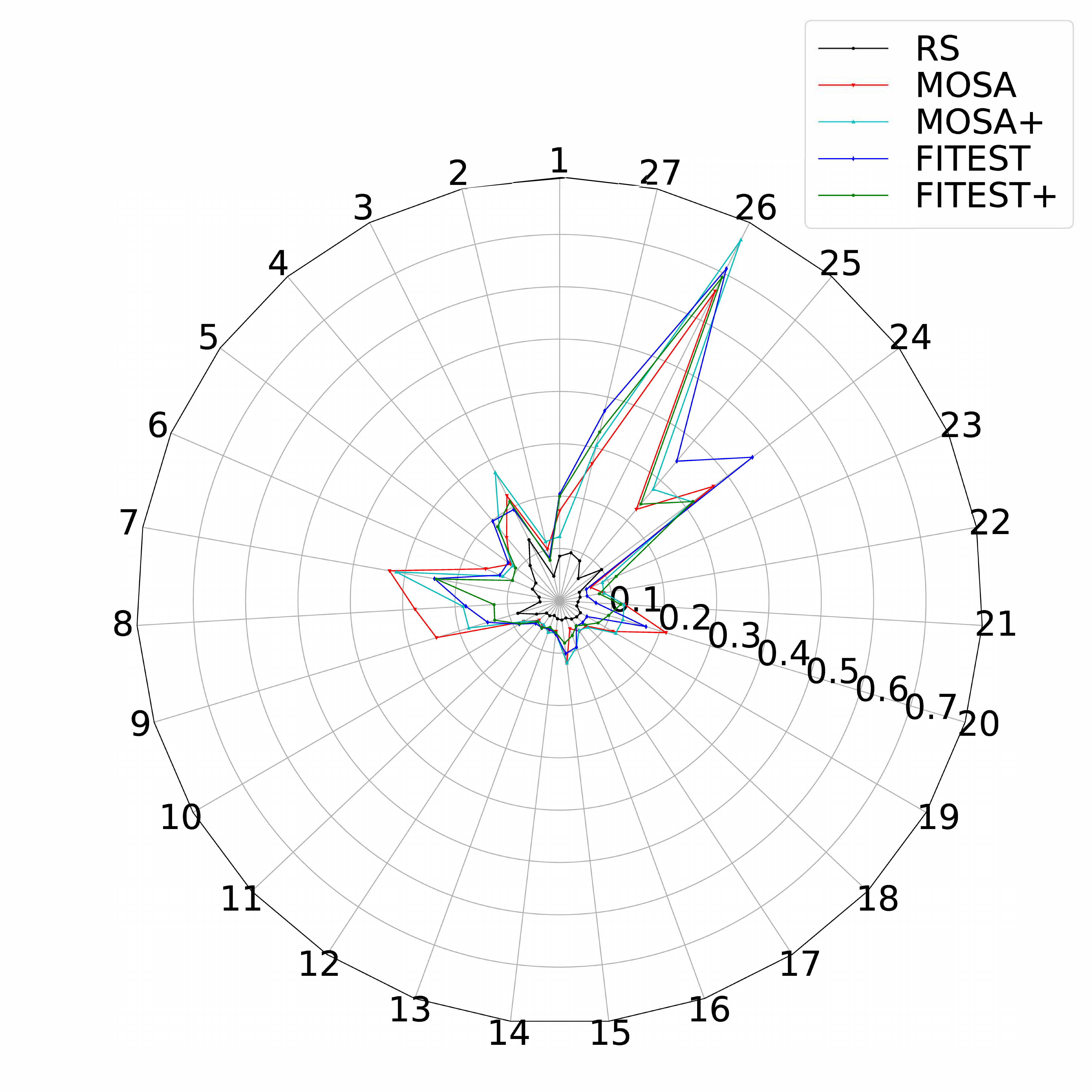}
	\caption{Misprediction Severity (\textit{MS}) for
	individual key-points for different search algorithms}
	\label{fig:radar}
\end{figure}

Figure~\ref{fig:radar} shows average \textit{MS} values of the test suites
generated by different search algorithms, for individual key-points.
For example, we can see that the average \textit{MS} value with MOSA+
for the 26th key-point (i.e., KP26) is around 0.8.

Comparing MOSA, FITEST, and their variants, we can see that the
overall patterns shown in the radar chart are similar. This implies
that the many-objective test suite generation algorithms yield similar
patterns in terms of which individual key-points tend to get higher \textit{MS}
values. To further assess the difference between them, we need to
check the results of the statistical tests.

Table~\ref{table:statistics} shows the results of statistical
comparisons between different search algorithms. Under the \textit{MS}
column, sub-columns \textit{$p$-value} and $\hat{A}_{AB}$ indicate the
statistical significance and the effect size, respectively, when
comparing the \textit{MS} distributions of two algorithms under columns
\textit{A} and \textit{B}.

In Table~\ref{table:statistics}, with $\alpha = 0.01$, there is no statistically
significant difference in \textit{MS} between MOSA and MOSA+, and between FITEST
and FITEST+. This implies that, consistent with RQ1, dynamically adjusting the
distribution index in crossover does not increase misprediction severity for
individual key-points.

As for MOSA and FITEST, which were significantly different with respect to
\textit{ES} in RQ1, they are no longer significantly different regarding
\textit{MS}. This inconsistency between RQ1 and RQ2 happens because, even though
MOSA (and MOSA+) is better than FITEST (and FITEST+) in making the $\mathit{NE}$
values of more key-points exceed $\epsilon = 0.05$, FITEST (and FITEST+) yields
higher $\mathit{NE}$ values than MOSA (and MOSA+) for some key-points. When it
comes to MOSA+ and FITEST+,  MOSA+ is significantly better than FITEST+ with a
small effect size (0.56). The results between MOSA and FITEST, and MOSA+ and
FITEST+, imply that dynamically reducing the population size during the search
increases the degree of mispredictions for some key-points, but the overall
impact is limited.

Interestingly, Figure~\ref{fig:radar} shows that some key-points (i.e., KP7,
KP24, KP25, KP26, and KP27) are more severely mispredicted than the others. A
detailed analysis found two distinct reasons that together affect the accuracy
of the DNN: the under-representation of some key-points in the training data and
a large variation in the shape and size of the mouth across different 3D models.
For KP7, we found that it is only included in 79\% of the training data.  One
possible explanation is that, as shown in Figure~\ref{fig:27-keypoints}, KP7 can
easily become invisible depending on the head posture. Interestingly, KP9, which
has a symmetrical position to KP7 in the face, is included in a larger
proportion of the training data (84\%). Such situations typically happen when the
training set is not built in a systematic fashion. This implies that key-points
that are under-represented in training data, because they happen not to be
visible on numerous occasions, are more likely to be severely mispredicted. On
the other hand, KP24, KP25, KP26, and KP27 are more severely mispredicted than
the others, even though they are included in most of the training images (more
than 92\%). These four key-points are  severely mispredicted because they are
related to the mouth, which shows the largest variation among face features; the
mouth can be opened and closed, and is larger than the eyes. Therefore, the
positions of the mouth key-points vary more than other key-points in the face,
even when the head posture is fixed. As a result, learning the position of the
mouth key-points is more difficult than that for other key-points. This implies
that, during training, we need to focus more on certain key-points whose actual
positions can vary more than the others across images.

To summarize, the answer to RQ2 is that, by additionally considering \textit{MS}, we
cannot further distinguish MOSA and FITEST, and their variants. Instead, through
a detailed analysis of the most severely mispredicted key-points, we identified
important insights to prevent severe mispredictions: (1) since some key-points
may not actually be visible in an image, depending on image characteristics
(e.g., head posture), for each key-point, it is important to ensure that the
training data contains enough images where the key-point is visible, (2) more
training is required for certain key-points, whose positions tend to vary
significantly more across the face and which are harder to predict than the
others. Such observations can obviously generalize to other types of images and
key-points.

\subsection{RQ3: Explaining Mispredictions}\label{sec:rq3}

\subsubsection{Setup}\label{sec:rq3-setup}
To answer RQ3, decision trees~\cite{breiman1984classification} are
learnt to infer how the $\mathit{NE}$ (normalized prediction error) of
the FKP-DNN for individual key-points relate to image characteristics
used by the simulator to generate test images. We use decision trees
because they are easy to interpret due to their hierarchical
decision-making process~\cite{murdoch2019definitions}, though
alternative forms of rule-based learning could be considered as well.
Interpretability is essential for engineers to assess the
risks associated with a DNN in the context of a given application and
to devise ways to improve the DNN, for example, through additional
training data. Specifically, we build a decision tree for each
key-point to identify conditions describing how $\mathit{NE}$ varies
according to input variables, i.e., roll, pitch, yaw, and
3D model ID. Since our test suite generation approach generates many
test images specified by roll, pitch, yaw, and 3D model, and given
that $\mathit{NE}$ values are calculated for individual key-points
during the search process, we can use this information as a set of
observations for building decision trees and explain mispredictions.
To mimic a practical scenario in which engineers use our approach
overnight (e.g., ten hours), we use the information collected from
five random runs (i.e., equivalent to ten hours) of MOSA+ (i.e., the
most effective search algorithm according to RQ1 and RQ2). Since the
target variable (i.e., $\mathit{NE}$) is continuous, we use regression
trees to explain and predict the degree of mispredictions.
Specifically, we use REPTree (i.e., fast tree learner using
information-gain and reduced-error pruning) implemented in
Weka~\cite{witten2002data}.

To evaluate the (predictive) accuracy of generated regression trees,
we measure the Mean Absolute Error (MAE) using 10-fold cross validation.
Note that, to provide interpretable explanations, the simplicity of
the resulting regression trees is just as important as its accuracy. In
particular, as the number of nodes in a regression tree increases, it
becomes increasingly difficult for engineers to interpret the results.
As a trade-off between accuracy and simplicity, we set the minimum
number of observations per leaf node to 40, based on preliminary
experiments. For the other parameters for REPTree, we use default
values provided by Weka.

\subsubsection{Results}\label{sec:rq3-results}

\begin{figure}
\centering
\subfloat[Error\label{fig:tree:MAE}]
{\includegraphics[height=4cm]{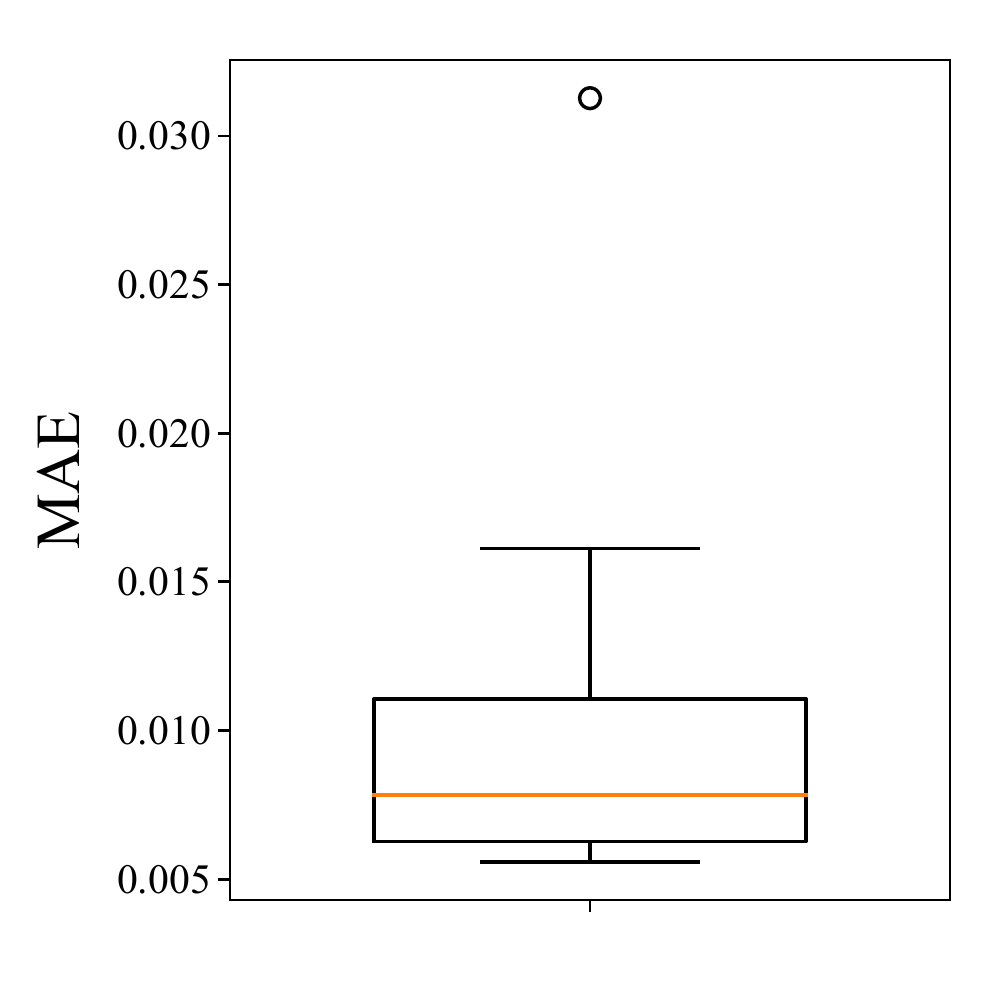}}
\hfill
\subfloat[Size\label{fig:tree:size}]
{\includegraphics[height=4cm]{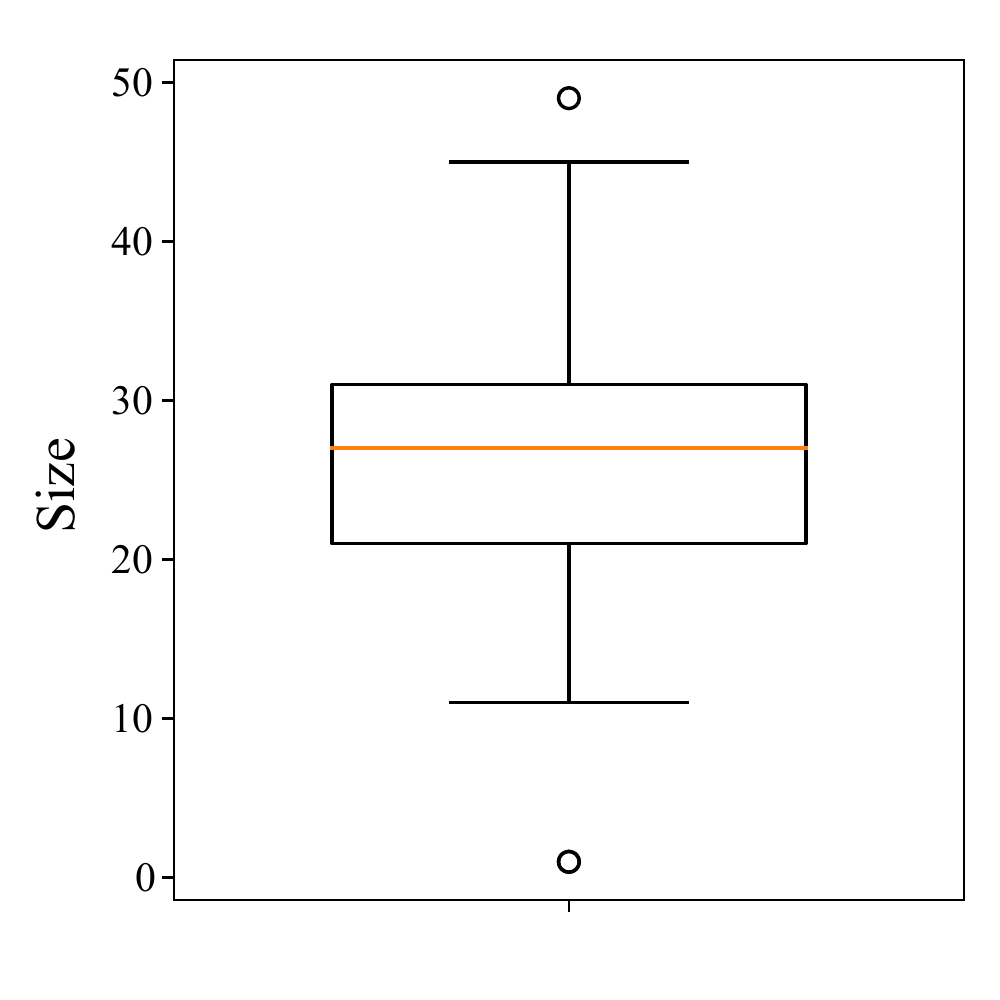}}
\caption{Regression tree accuracy and size for all key-points}
\label{fig:tree-summary}
\end{figure}

Figure~\ref{fig:tree-summary} shows the MAE and size (i.e., number of
nodes) distributions for all 27 trees built based
on 3854 test images obtained from five runs of MOSA+. The average MAE
is 0.01 and the average size is 25.7. We can see that the trees are
accurate as there is an average difference of 0.01 between the actual
and predicted $\mathit{NE}$ values, which is far
below the threshold that IEE finds acceptable (0.05). The trees are
also of reasonable size and therefore easy to interpret as they lead
to simple rules, as further shown below.

Table~\ref{tab:KP26-rules} shows some representative rules derived
from the tree generated for KP26, the most severely mispredicted
key-point in RQ2. All the remaining rules and the trees for the other
key-points are available in the supporting
material~\footnote{All the decision trees are available on https://figshare.com/s/0433751b88282b07ea12. Unfortunately, the full replication package could not be made publicly available since the IEE-SIM and the IEE-DNN are proprietary assets.}.
In Table~\ref{tab:KP26-rules}, column \textit{$\mathbf{ic}$-condition}
shows conditions on values of the image characteristics, i.e., Roll (R),
Pitch (P), Yaw (Y), and 3D Model ID (M); column $\mathit{NE}$ shows
the average $\mathit{NE}$ value for all test images satisfying the
condition. For example, the first row from the table means that, for
test images such that 3D model ID is 9 and pitch is less than 18.41,
the average $\mathit{NE}$ is 0.04. Recall that each 3D model ID
corresponds to implicit facial characteristics. For example, for
model ID 9, skin color is brown, face structure is broad, nose type is
aquiline, and mouth structure is uneven.

\begin{table}
    \centering
    \caption{Representative rules derived from the decision tree for KP26
    (M: Model-ID, P: Pitch, R: Roll, Y: Yaw)}
    \begin{tabularx}{\linewidth}{Xl}
    \toprule
    $\mathbf{ic}$-condition & $\mathit{NE}$ \\
    \midrule
    $M=9 \wedge P<18.41$ & $0.04$ \\
    $M=9 \wedge P \ge 18.41 \wedge R < -22.31 \wedge Y < 17.06$ & $0.26$ \\
    $M=9 \wedge P \ge 18.41 \wedge R < -22.31 \wedge 17.06 \le Y < 19$ & $0.71$ \\
    $M=9 \wedge P \ge 18.41 \wedge R < -22.31 \wedge Y \ge 19$ & $0.36$ \\
    \bottomrule
    \end{tabularx}
    \label{tab:KP26-rules}
\end{table}

\begin{figure}
\centering
\subfloat[A test image satisfying the first condition in Table~\ref{tab:KP26-rules}\label{fig:rq3:small-error}]
{\includegraphics[height=3.7cm]{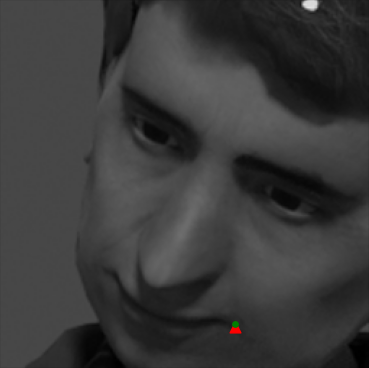}}
\hspace{5mm}%
\subfloat[A test image satisfying the third condition in Table~\ref{tab:KP26-rules}\label{fig:rq3:large-error}]
{\includegraphics[height=3.7cm]{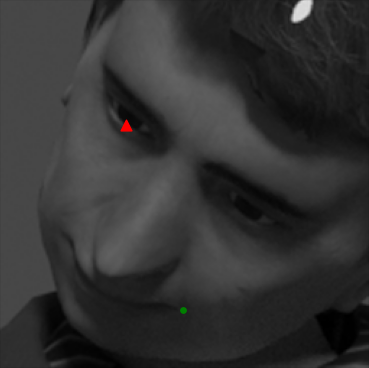}}
\caption{Test images satisfying different conditions
in the regression tree for KP26. Actual and predicted positions
are shown by green-circle and red-triangle dots, respectively.}
\label{fig:rq3}
\end{figure}

Using such conditions, engineers can easily identify when the FKP-DNN leads to
severe mispredictions, for individual key-points. For example, by comparing the
first and third conditions in Table~\ref{tab:KP26-rules}, we can see that, for
the same 3D model, changing the head posture toward specific ranges leads to a
significant increase of the prediction error of the IEE-DNN for KP26. To
better visualize the example above, we present two test images, in
Figures~\ref{fig:rq3:small-error} and \ref{fig:rq3:large-error}, satisfying the
first and third conditions, respectively. In each image, the green dots and red
triangle indicate the actual and predicted positions of KP26. While there is a
small prediction error ($\mathit{NE}=0.013$) in
Figure~\ref{fig:rq3:small-error}, it gets much larger ($\mathit{NE}=0.89$) when
turning the head further down and to the right, as shown in
Figure~\ref{fig:rq3:large-error}. Engineers can further investigate the root
causes of severe mispredictions using targeted additional images. For example,
based on Figure~\ref{fig:rq3:large-error}, we generated a targeted image that
differs only in the direction of the shadow (by changing the position of the
light source in the 3D model) and could confirm that the shadow was indeed the
root cause of the large $\mathit{NE}$ value.

Knowing under what conditions severe mispredictions are occurring can
help engineers in two ways. First, it helps to assess the risks
associated with individual key-points for specific conditions, in the
context of a specific application. For example, if the head posture
can be constrained to $P < 18.41$ in the context of a certain
application of the FKP-DNN, the risk of severe misprediction for KP26
is reduced according to the conditions in Table~\ref{tab:KP26-rules}.
Second, it enables the generation of specific test images, using the
simulator, that are expected to cause particularly severe
mispredictions and can be used for retraining the DNN. For
example, from the conditions in Table~\ref{tab:KP26-rules}, we can
generate new images satisfying $M=9 \wedge P \ge 18.41 \wedge R <
-22.31$, which will likely lead to high $\mathit{NE}$ values, and can
be used to help retrain and improve the DNN by augmenting its training
data set.

To summarize, the answer to RQ3 is that, by building regression trees
using the information that is produced by executing our test
generation approach, we can largely explain the variance in
individual key-point mispredictions in terms of image
characteristics that are controllable in the simulator. These
conditions can help engineers assess risks for individual key-points
and generate specific images for retraining of the DNN in a way that
is more likely to have an impact on its accuracy.

\subsection{Threats to Validity}
In RQ1, we have used a threshold ($\epsilon = 0.05$) for deciding if a
key-point is severely mispredicted or not. This may affect the
results. However, this threshold value was set by IEE
domain experts, accounting for the specific application of their DNN.
Furthermore, we checked that using other small threshold values does not severely
affect overall trends but only slightly affect ES values.

We have used 3D face models with various facial characteristics
shared by IEE. IEE engineers have prepared the 3D models by marking
key-points' actual positions manually. The manual marking
can be erroneous and such error can affect the
quality of data generated for training and testing the DNN. To
mitigate such threats, IEE engineers have done extensive testing for
each 3D model and carefully validated the generated data.

Though we focused, in our case study, on one FKP-DNN developed by IEE, it is
representative of what one can find in
industry~\cite{huang2015deepfinger,chen2017recurrent,jabbar2018real},  in terms
of prediction accuracy, inputs and outputs, and training procedure.
Though there are no publicly available KP-DNNs coupled with a simulator, additional
industrial case studies are, however, required to improve the generalizability
of our results.

Since we focused on syntactic images generated by a simulator, the
transferability of our results from simulation to real-world facial
key-points detection is not in the scope of this work. Nevertheless,
we want to note that \citet{9159088} already provided an experimental
methodology to quantify such transferability and reported that
simulator-generated data can be a reliable substitute to real-world
data for the purpose of DNN testing.

\subsection{Lessons Learned}

\subsubsection*{\textbf{Lesson 1: Automated test suite generation is indeed useful in practice.}}
Although we have presented here the evaluation results using the latest
IEE-DNN version, our approach has been continuously applied to previous
versions during its development process. Indeed, test suites automatically
generated by our approach and the analysis of testing results for mispredicted
key-points have helped assess and improve IEE-DNNs. In particular, our
approach led IEE to augment their training dataset (e.g., diversity and
sample size) and thus improve the generalization of the IEE-DNN against
various inputs. In summary, based on the results, IEE has been (1)
continuously enriching their dataset by adding more training images from diverse
3D face models and (2) improving the IEE-DNN's architecture by doubling
the number of hidden layers to drastically increase its accuracy.

Our approach has been also used for improving the simulator. For
example, during a detailed analysis of the testing results, we
revealed a critical issue: the labeled key-point positions on test
images, generated by the IEE-SIM, were not accurate and, after
analysis, engineers realized there was a misalignment between the
texture and nodes of the 3D mesh that are used to define the actual
position of key-points.

IEE also plans to use our approach in various ways. Since the test suites
generated by our approach cause severe mispredictions for many key-points,
retraining the DNN using these test suites---given that the DNN architecture is
now adequate---is expected to help improve the accuracy of the DNN.
We will further investigate to what extent such accuracy can be improved
in our future work.

\subsubsection*{\textbf{Lesson 2: Understanding mispredictions is critical.}}
Though explaining DNNs~\cite{samek2017explainable} is important in many circumstances, the complex structure of such models makes it challenging. Instead of analyzing the internals of the IEE-DNN,
we simply built decision trees for inferring conditions characterizing
key-point mispredictions in terms of input variables, i.e., roll,
pitch, yaw, and 3D model ID. Even such analysis cannot directly
explain the behavior of the IEE-DNN in detail, knowing such conditions
brings several benefits to IEE. First, it helps them demonstrate the
robustness of the DNN under certain conditions which, for generated
test images, do not lead to severe mispredictions. If those conditions
match the conditions of the intended application, then engineers can
conclude that the DNN is likely to be safe for use. Second, it facilitates the investigation of the root causes of
mispredictions. This is critical because, similar to the severe
misprediction in Figure~\ref{fig:rq3:large-error} discussed in
\S~\ref{sec:rq3-results}, some mispredictions are not avoidable and
can only be addressed by engineering the system
using the DNN for robustness (e.g., by identifying shadowed areas and not using
predictions for the key-points in these areas). Such finding
led IEE to better target their development resources to improve the driver's gaze detection system rather than just focusing on the IEE-DNN itself.

\subsubsection*{\textbf{Lesson 3: Simulation-based testing brings key benefits.}}
As discussed in Section~\ref{sec:sim}, a simulator is one of the key
enablers of our approach as it generates \textit{labeled} test images
in a controlled manner. However, the simulator can also be a limiter as
test suites generated by our approach depend on what the simulator can generate.
Ideally, one would like access to a high-fidelity and configurable simulator
that is able to generate diverse images, accounting for all major factors
that affect the behavior of the DNN under test. For example, the
IEE-SIM has been developed to support the configuration of various
image characteristics, such as head posture, face size, and skin
color, that are essential because of their effect on facial key-point
detection. Thanks to the simulator, we can generate as many different
test images as we need, with known key-points, and drive effectively
automated test generation. Given the usefulness of our approach, the
cost of having a simulator, with sufficient fidelity and configurability,
benefits the development and testing of KP-DNNs by enabling efficient automation.

\section{Conclusion} \label{sec:conclusion}
In this paper, we formalize the problem definition of KP-DNN testing and present an approach to automatically generate test data for KP-DNNs with many independent outputs, a common situation in many applications. 
We empirically compare state-of-the-art, many-objective search algorithms and their variants tailored for test suite generation. 
We find MOSA+ to be significantly more effective than random search (baseline) and other many-objective search algorithms, e.g., FITEST, with large effect sizes. 
We also observe that our approach can generate test suites to severely mispredict more than 93\% of all key-points on average, while random search, as a comparison, can do so for 41\% of them.
We further investigate and demonstrate a way, based on regression trees, to learn the conditions, in terms of image characteristics, that cause severe mispredictions for individual key-points. These conditions are essential to engineers to assess the risks associated with using a DNN and to generate new images for DNN retraining, when possible.

As a part of future work, we plan to devise effective retraining strategies and assess to what extent the accuracy of FKP-DNNs can be improved through retraining. We also plan to tailor NSGA-III, a generic many-objective search algorithm, for test suite generation and empirically evaluate its performance.

\begin{acks}
This work has received funding from Luxembourg’s National Research Fund (FNR) under grant BRIDGES2020/IS/14711346/FUNTASY, the European Research Council under the European Union’s Horizon 2020 research and innovation programme (grant agreement No 694277), IEE S.A. Luxembourg, and NSERC of Canada under the Discovery and CRC programs.
Donghwan Shin was partially supported by the Basic Science Research Programme through
the National Research Foundation of Korea (NRF) funded by the Ministry of Education (2019R1A6A3A03033444).
\end{acks}

\bibliographystyle{ACM-Reference-Format}
\bibliography{DNN-Testing-bibliography}

\end{document}